\documentclass{article}
\usepackage{paperpilot}

\title{Image Editing Models are Numerical Simulators}
\author{Ulysse Mizrahi\\Tel Aviv University}
\date{}

\begin{document}
\maketitle

\begin{figure}[htbp]
  \centering
  \includegraphics[width=\linewidth]{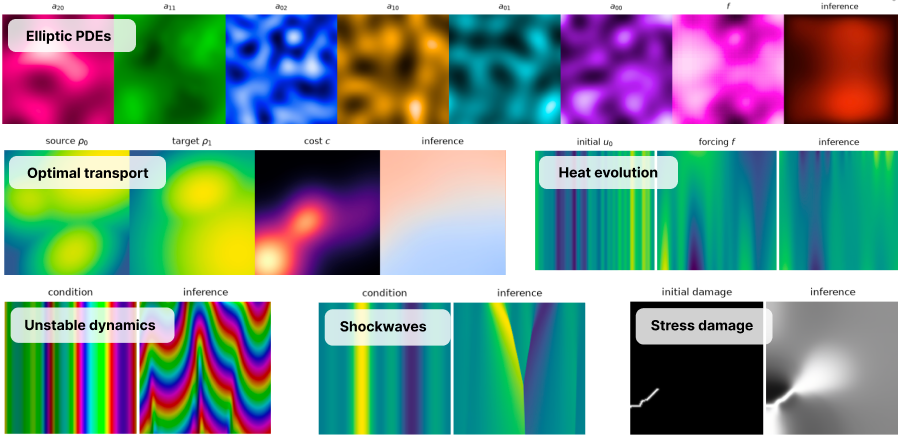}
\end{figure}

\begin{abstract}
We investigate whether a pretrained generative image-editing model can provide a common interface for numerical simulation. Physical inputs and solutions are rendered as images, while scalar quantities such as material properties, diffusivity, and loading parameters enter through lightweight adapters. Using established numerical and analytic solvers for supervision, we apply the same architecture and training protocol to heterogeneous elliptic equations, forced heat and Burgers evolution, complex Ginzburg-Landau dynamics, two-dimensional Navier-Stokes prediction, potential flow, elasticity, eikonal travel time, phase-field fracture, and entropic optimal transport. The results show that a pretrained image model can represent diverse static and time-dependent physical mappings, including unstable and shock-like behavior, when each task is expressed through a suitable visual encoding. This work is a capability study rather than an attempt to surpass specialized solvers. It also identifies fundamental constraints: image and latent representations complicate numerical range selection and direct enforcement of governing equations or invariants, while a failed Kuramoto-Sivashinsky experiment indicates that representation errors prevent meaningful long-horizon simulation of chaotic systems.
\end{abstract}

\section{Introduction}
Numerical simulation is usually organized around specialized discretizations, solvers, and data structures. Although these tools are accurate and well understood, each system brings its own representation of geometry, coefficients, boundary conditions, state variables, and outputs. A one-dimensional evolution equation, a two-dimensional elasticity problem, and an optimal-transport problem may all produce spatial fields, yet their distinct dynamics call for different numerical treatments. Chaotic systems are particularly demanding: a positive Lyapunov exponent quantifies exponential sensitivity to perturbations, so discretization and representation errors limit the horizon over which two trajectories remain close \cite{eckmann1985ergodic}.

Many PDE and physics problems are nevertheless naturally image-like. Heterogeneous elliptic equations contain coefficient and forcing fields; time-dependent PDEs form space-time diagrams; and cross-sections can represent aerodynamics, elasticity, fracture, wave propagation, and transport. This observation suggests a common interface in which the physical instance is rendered as one or more conditioning images and the corresponding solution is rendered as a target image.

Image-editing models are designed primarily for semantic or aesthetic transformations of natural images, such as replacing objects, changing style, or modifying visual attributes from reference images and text instructions. Yet modern editors have demonstrated substantial generalization beyond narrowly specified transformations, and their conditioning mechanisms can be adapted with relatively small trainable modules. Our central hypothesis is that an editor can be fine-tuned to map visual encodings of physical inputs to visual encodings of numerical solutions. Lightweight conditioning modules can additionally expose scalar material or equation parameters that are not conveniently represented as images.

We test this hypothesis with a FLUX-family image-editing backbone \cite{blackforestlabs2025kontext}, low-rank adaptation (LoRA) \cite{hu2022lora}, and, where needed, adaptive layer-normalization conditioning initialized at zero (AdaLN-Zero) \cite{peebles2023dit}. The model receives physical images and optional normalized parameters and generates solution fields at $256\times256$ resolution. Standard analytic and numerical methods provide supervision. The resulting interface is evaluated on static fields and space-time renderings spanning elliptic PDEs, one-dimensional evolution equations, two-dimensional Navier-Stokes evolution, potential flow, elasticity, eikonal wave propagation, phase-field fracture, and entropic optimal transport.

Our objective is not to replace specialized solvers or claim better accuracy than established numerical methods. Instead, the experiments ask how broadly a pretrained image editor can represent physical mappings once their inputs and outputs are expressed through its native visual interface. The study is therefore a capability demonstration that also exposes important limitations in accuracy, conservation, stability, and chaotic dynamics.

\section{Related Work}
\subsection{Neural simulation}
Neural simulation methods learn solution operators or time-stepping rules from data. They can amortize expensive solves, interpolate across parameter settings, and provide differentiable surrogates.

\paragraph{Physics-informed neural networks.}
Physics-informed neural networks (PINNs) \cite{raissi2019pinn} represent a solution as a neural function of continuous space-time coordinates. Automatic differentiation supplies the derivatives needed to penalize the governing equation together with boundary, initial-condition, and observation errors. This direct use of the equation is useful when measurements are sparse or paired numerical solutions are unavailable, particularly for inverse problems and data assimilation. Optimization can nevertheless be difficult: the composite losses can produce severely unbalanced back-propagated gradients and stiff training dynamics unless their contributions are balanced \cite{wang2021gradientpathologies}. A new optimization problem is also typically solved for each equation instance or parameter setting.

\paragraph{Neural operators.}
Neural operators learn maps between functions, such as the map from a coefficient field, forcing term, or initial condition to a solution field. DeepONet \cite{lu2021deeponet} evaluates this operator by combining a branch network that encodes samples of the input function with a trunk network that encodes output coordinates. The Fourier neural operator (FNO) \cite{li2021fno} instead learns global convolution kernels in Fourier space and alternates them with pointwise transformations. Once trained, either architecture can amortize an entire family of PDE solves. The operator formulation can be evaluated on discretizations different from those used during training, and FNO demonstrated zero-shot super-resolution. The learned map nevertheless covers the sampled operator family; new equations, boundary conditions, geometries, or parameter regimes are not handled automatically \cite{kovachki2023neuraloperator}.

\paragraph{Learned discretizations and time steppers.}
PDE-Net \cite{long2018pdenet} learns convolutional filters constrained to approximate differential operators and combines them with learned nonlinear response functions. Local filters resemble finite-difference stencils, while repeated layers predict the evolution of gridded fields and can expose symbolic or semi-symbolic equation structure. The method is designed to discover local evolution laws from regularly sampled trajectories rather than to solve arbitrary boundary-value problems or changing geometries.

\paragraph{Graph and mesh neural simulators.}
Graph Network-based Simulators and MeshGraphNets \cite{sanchezgonzalez2020learning,pfaff2021meshgraphnets} represent particles or mesh elements as graph nodes and propagate local interactions along edges through message passing. Their topology accommodates particles, deformable bodies, adaptive meshes, and complex geometries while scaling to different graph sizes. Both learn autoregressive update rules, so rollout quality depends on controlling accumulated one-step error; Graph Network-based Simulators explicitly perturb training states with noise for this purpose. Message passing alone does not impose exact conservation or stability.

\subsection{Generative image editing}
Generative image editors transform one or more input images, often under a text instruction, into a modified image.

\paragraph{GAN-based image translation and latent editing.}
Generative adversarial networks (GANs) \cite{goodfellow2014gan} train a generator to produce samples that a discriminator cannot distinguish from real data. The conditional GAN pix2pix \cite{isola2017pix2pix} applies this adversarial objective to paired image-to-image examples and supplements it with a reconstruction loss. CycleGAN \cite{zhu2017cyclegan} removes the need for aligned pairs by learning mappings in both directions and requiring a round trip to reconstruct the original image. These systems established practical image translation with fast inference, but pix2pix requires paired examples and CycleGAN substitutes a cycle-consistency assumption. Conventional adversarial training can also be unstable and suffer mode collapse, in which the generator covers too little of the data distribution \cite{arjovsky2017wgan}.

Later work moved editing into structured generator latents. StyleGAN \cite{karras2019stylegan} maps random inputs through an intermediate style space that controls synthesis at multiple resolutions; InterFaceGAN \cite{shen2020interfacegan} finds linear directions in that space for semantic attributes; and StyleCLIP \cite{patashnik2021styleclip} uses CLIP guidance to connect text instructions to StyleGAN edits. These representations permit interactive control and high visual quality within the generator's domain, but real-image inversion and out-of-domain edits trade reconstruction fidelity against editability \cite{tov2021e4e}.

\paragraph{Early diffusion editing.}
Diffusion and score-based models \cite{ho2020ddpm,song2021scorebasedsde} learn to reverse a gradual noising process through iterative denoising. SDEdit \cite{meng2021sdedit} perturbs an input image to an intermediate noise level and denoises it under a generative prior, trading source preservation against realism. RePaint \cite{lugmayr2022repaint} performs inpainting by repeatedly restoring known pixels during reverse diffusion, whereas Blended Diffusion \cite{avrahami2022blendeddiffusion} combines a spatial mask with CLIP guidance for local text-driven edits. Latent diffusion models \cite{rombach2022ldm} reduce the cost of high-resolution generation by denoising a compressed autoencoder representation and injecting conditions through cross-attention.

Several subsequent methods target more precise controls. Prompt-to-Prompt \cite{hertz2022prompttoprompt} reuses and modifies cross-attention maps to preserve layout under prompt changes; InstructPix2Pix \cite{brooks2023instructpix2pix} trains directly on instruction-following image pairs; and Paint by Example \cite{yang2022paintbyexample} encodes a reference image as the edit condition. Iterative sampling remains comparatively slow, but few-step distillation can reduce the number of model evaluations. Progressive Distillation \cite{salimans2022progressive} repeatedly trains a student to reproduce two teacher steps with one step, halving the sampling trajectory at each stage. Consistency Models \cite{song2023consistency} instead learn to map points on the same probability-flow trajectory to a common clean prediction, permitting one- or few-step generation. These acceleration methods reduce sampler cost, but they do not add numerical constraints to the editing objective.

\paragraph{FLUX-like flow and transformer editors.}
Recent image generators combine compressed latent spaces, transformer backbones, and flow-based objectives. Flow Matching \cite{lipman2023flowmatching} trains continuous normalizing flows with a simulation-free objective that regresses vector fields of fixed conditional probability paths. Rectified Flow \cite{liu2022rectifiedflow} learns straighter transport paths between noise and data so that they can be integrated with fewer steps, and rectified-flow transformers scale this formulation to high-resolution text-to-image synthesis \cite{esser2024rectifiedflow}. The FLUX family follows this design: FLUX.1 \cite{blackforestlabs2024flux1dev} is a text-to-image model, while FLUX.1 Kontext \cite{blackforestlabs2025kontext} extends the transformer to in-context generation and editing from text and image inputs. Rich image conditioning makes this family attractive for physical mappings, but its latent representation remains an image prior rather than a numerically lossless field representation. Physical validity therefore depends on encoding design, bounded value ranges, parameter injection, and the adaptation procedure.

\subsection{Generative models for simulation}

\paragraph{Generative simulators trained on physical data.}
Generative simulators trained directly on physical data use learned distributions to represent uncertainty, infer missing fields, or solve forward and inverse problems from partial observations. Physics-informed diffusion for flow reconstruction \cite{shu2023physicsinformeddiffusion} uses a diffusion prior to reconstruct high-fidelity flow fields and can incorporate PDE-derived conditioning when the governing equation is known. DiffusionPDE \cite{huang2024diffusionpde} learns a joint distribution over PDE coefficients and solutions, so the same conditional sampler can perform forward or inverse inference with incomplete observations. VideoPDE \cite{li2025videopde} casts spatiotemporal PDE solving as video inpainting: a transformer diffusion model fills unknown space-time pixels around arbitrary observed subsets.

Pixel-space diffusion avoids a pretrained natural-image codec, while explicit observation masks and PDE-derived conditioning preserve the information these methods are designed to use. The tradeoff is task-specific simulation data and iterative sampling.

\paragraph{Pretrained generative backbones for simulation.}
A newer branch adapts large pretrained visual generators as physical surrogates. WinDiNet \cite{perini2026windinet} fine-tunes the LTX-Video latent transformer on urban CFD frames, transferring spatial and temporal video priors to wind prediction while retaining differentiability for gradient-based layout optimization. PhysiX \cite{nguyen2025physix} initializes a discrete tokenizer and a 4.5B-parameter autoregressive transformer from video-generation checkpoints, jointly trains across several physics datasets, and applies a refinement module to reduce tokenization artifacts in continuous fields. These studies indicate that visual pretraining can supply useful priors for motion and long-range dependence even when channels encode velocity, pressure, or temperature rather than natural RGB appearance.

The central difficulty for pretrained visual backbones is the domain gap. The cited methods introduce tokenizer or VAE adaptation and field-refinement stages to recover numerical accuracy from representations developed for natural imagery. Their outputs must therefore be evaluated in physical units rather than judged by visual plausibility alone.

\section{Method and Results}
\subsection{General approach}
Each experiment defines a task-specific but deterministic visual encoding of its physical inputs and solution. Scalar or vector fields are mapped to RGB images with documented numerical ranges and color conventions; multiple input fields become separate conditioning images, while time-dependent one-dimensional fields are represented as space-time images. The inverse mapping converts the generated pixels back to physical quantities when numerical comparison is required. This interface preserves the native structure of each task while presenting every experiment to the generative model as image editing.

We use FLUX.2-klein-4B \cite{blackforestlabs2026flux2klein4b} as the common backbone. Its frozen variational autoencoder (VAE) \cite{kingma2014vae} maps the condition and target images to latent tokens. During training, condition tokens are concatenated with noisy target tokens, and the transformer learns to reconstruct the target under the same image-conditioning interface used by the pretrained editor. The current implementation fine-tunes a separate adapter set for each equation rather than training a single checkpoint across all tasks.

Low-rank adaptation (LoRA) \cite{hu2022lora} is applied to the query, key, value, and output projections of dual-stream attention and to the fused query-key-value/MLP projections of single-stream blocks. Additional output-projection adapters modify the first 24 single-stream blocks. When image conditioning does not fully specify a problem, normalized scalar physical parameters are injected into every transformer block through AdaLN-Zero \cite{peebles2023dit}. Each dual-stream block receives six modulation vectors: shifts, scales, and residual gates for its attention and MLP sublayers. This separates spatially structured inputs from low-dimensional controls such as diffusivity, material constants, or applied loads.

The trainable adapters and conditioning projections are optimized by conditional flow matching \cite{lipman2023flowmatching} on the backbone's four-step distilled schedule. Few-step distillation methods show how a long generative trajectory can be compressed into a small number of evaluations \cite{salimans2022progressive,song2023consistency}; here we retain the schedule supplied with the already distilled base model rather than distilling it again. Appendix~\ref{app:model-implementation} gives the modulation equations, flow-matching objective, and single-stream adaptation details.

\subsection{Many image conditionings: 2D Elliptic PDEs}
\paragraph{Problem description.}
Elliptic boundary-value problems generalize familiar equilibrium equations such as the Laplace and Poisson equations; examples include steady heat conduction, electrostatic potential, and groundwater hydraulic head \cite{narasimhan2008laplace,leveque2007finite}. The task is to infer a scalar solution from six coefficient fields and a forcing field.

\paragraph{Equation formulation.}
On $\Omega=[0,1]^2$, the generator solves
\begin{equation}
 a_{20}u_{xx}+a_{11}u_{xy}+a_{02}u_{yy}
 +a_{10}u_x+a_{01}u_y+a_{00}u=f,
 \qquad u|_{\partial\Omega}=0.
\end{equation}
Here $u$ is the unknown field, $f$ is the forcing, subscripts denote partial derivatives, and the six coefficient fields weight the second-, first-, and zeroth-order terms. The generator samples $a_{20}>0$, $a_{02}>0$, and $a_{11}^2<4a_{20}a_{02}$, making the principal coefficient matrix positive definite and the operator elliptic.

\paragraph{Numerical solver method.}
A centered finite-difference discretization produces a nine-point stencil on a $256\times256$ grid, after which the sparse linear system is solved directly \cite{leveque2007finite}. Appendix~\ref{app:solver-elliptic} specifies the coefficient generator and stencil.

\paragraph{Image encoding.}
The six coefficient fields, forcing field, and solution are encoded as eight separate images using fixed, channel-specific numerical ranges and distinct color ramps.

\paragraph{In laymen's terms.}
The model receives seven colored maps describing how a material behaves and where it is pushed, then predicts the final balanced state once no further change is occurring.

\begin{figure}[htbp]
  \centering
  \includegraphics[width=\linewidth,height=0.72\textheight,keepaspectratio]{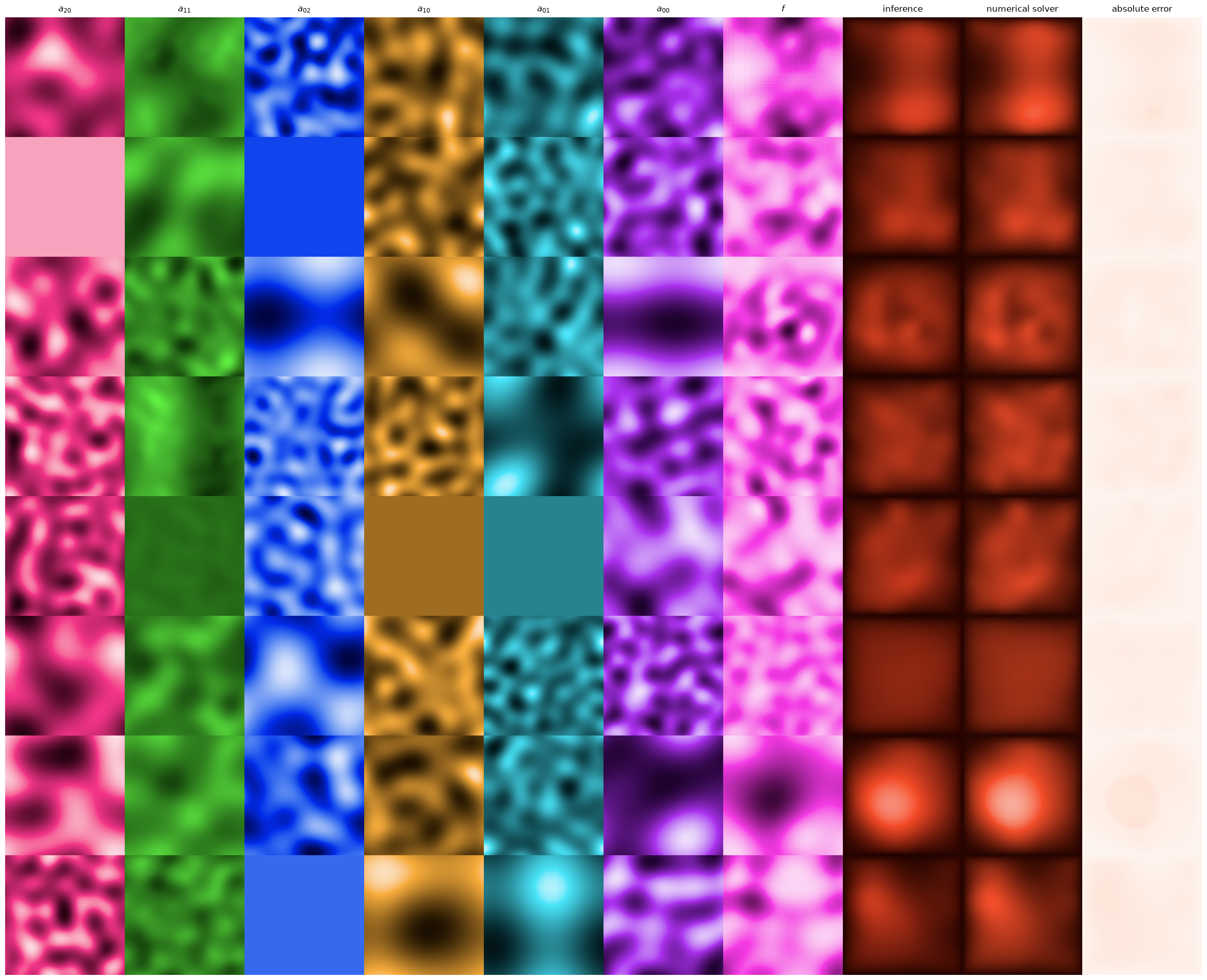}
  \caption{Random, non-handpicked held-out samples for the two-dimensional elliptic PDE. Each row shows the six coefficient channels $a_{20},a_{11},a_{02},a_{10},a_{01},a_{00}$, forcing $f$, model inference, numerical solution, and pointwise absolute error. Fixed channel-specific color ramps map each field's numerical values to color; inference and reference use the same solution ramp, while darker red indicates larger absolute error.}
  \label{fig:elliptic-results}
\end{figure}

\subsection{Parameter conditioning: 1D heat equation with forcing}
\paragraph{Problem description.}
The heat equation \cite{leveque2007finite} determines how temperature diffuses and responds to external heating or cooling. This task predicts a time-varying temperature field from an initial profile, a forcing field, and a thermal diffusivity. Although the PDE has one spatial dimension, its evolution forms a two-dimensional space-time image.

\paragraph{Equation formulation.}
For spatial coordinate $x\in[0,1]$ and time $t\in[0,T]$, the simulated equation is
\begin{equation}
  \partial_t u(x,t) = \kappa \partial_{xx}u(x,t) + q(x,t),
\end{equation}
with initial condition $u(x,0)=u_0(x)$. Here $u(x,t)$ is temperature, $\partial_t$ is the time derivative, $\partial_{xx}$ is the second spatial derivative, $\kappa>0$ is thermal diffusivity, $q(x,t)$ is the forcing term, and $T$ is the final simulated time. The spatial domain is periodic.

\paragraph{Numerical solver method.}
A Fourier analytic solver advances diffusion mode by mode with the exact heat multiplier and accumulates the forcing in the same spectral basis \cite{leveque2007finite}. Temporal and spatial sampling and random-field construction are detailed in Appendix~\ref{app:solver-heat}.

\paragraph{Image encoding.}
The horizontal image axis represents space and the vertical axis represents time. The initial condition is rendered as a time-constant image with each row equal to $u_0(x)$. The forcing field $q(x,t)$ and target temperature $u(x,t)$ are rendered as time-varying images under the same color scale. All three images share a symmetric per-sample color range determined by the largest magnitude in the forcing or solution.

\paragraph{Conditioning numerical parameters.}
The AdaLN-Zero scalar conditioning value for this method is the thermal diffusivity $\kappa$. Before normalization, $\kappa$ is transformed as $\log \kappa$, with $\kappa$ sampled between $10^{-5}$ and $10^{-2}$. This interval avoids both near-instant equilibration and negligible diffusion over the simulated horizon.

\paragraph{In laymen's terms.}
Given a starting temperature pattern, a changing source of heat, and a number describing how quickly heat spreads, the model draws the temperature at every later time.

\begin{figure}[htbp]
  \centering
  \includegraphics[width=\linewidth,height=0.76\textheight,keepaspectratio]{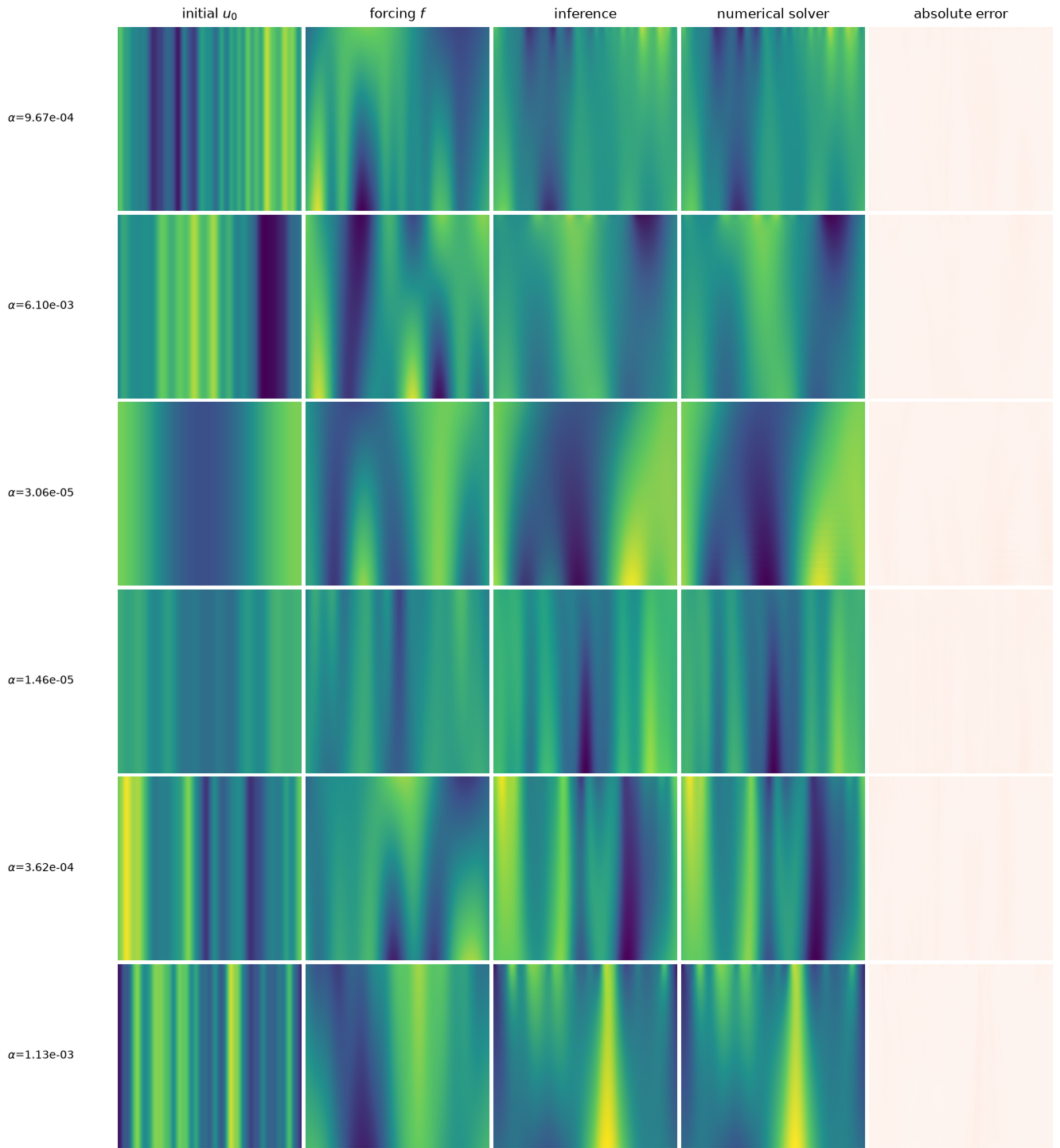}
  \caption{Random, non-handpicked held-out samples for the forced heat equation across diffusivities $\kappa$. Columns show initial condition $u_0$, forcing $q$, model inference, numerical solution, and pointwise absolute error. Purple denotes lower and yellow higher scalar values under a shared per-sample scale; darker red denotes larger error. The initial profile dominates early dynamics, while forcing becomes more visible later and diffusion smooths sharp gradients.}
  \label{fig:heat-results}
\end{figure}

\subsection{Unstable dynamics: 1D complex Ginzburg-Landau equation}
\paragraph{Problem description.}
The complex Ginzburg-Landau equation \cite{aranson2002world} evolves a one-dimensional complex field and can generate oscillatory waves, phase turbulence, and nonlinear saturation. Each state is a field $A(x,t)\in\mathbb{C}$. The equation serves as a reduced model for pattern formation near oscillatory instabilities in chemical reactions, nonlinear optics, lasers and cavities, and fluid wave trains.

\paragraph{Equation formulation.}
For spatial coordinate $x$ and time $t$, the simulated equation is
\begin{equation}
  \partial_t A(x,t)
  = A(x,t) + (1+i c_1)\partial_{xx}A(x,t)
  - (1-i c_3)|A(x,t)|^2A(x,t).
\end{equation}
Here $A(x,t)$ is the complex field, $i$ is the imaginary unit, $\partial_{xx}$ is the second spatial derivative, $c_1$ controls linear dispersion relative to diffusion, and $c_3$ controls the phase component of the cubic nonlinear saturation term. The simulation uses a periodic domain of length 128, $T=12$, $c_1=2$, and $c_3=1.2$.

\paragraph{Numerical solver method.}
A Fourier spectral discretization with an exponential-Euler integrator treats the linear diffusion-dispersion operator analytically and the cubic term explicitly \cite{cox2002etd}. Appendix~\ref{app:solver-cgl} provides the update equations, resolution, and initial-condition sampler.

\paragraph{Image encoding.}
The generated target is a space-time image whose horizontal axis is space and vertical axis is time. An HSV-like encoding maps complex phase to hue and magnitude to brightness. The condition image repeats the initial complex field along the time axis. Condition and solution share a per-sample magnitude scale to map to 0-1, and the simulated field is area-downsampled to $256\times256$ before colorization with saturation fixed at 0.95.

\paragraph{In laymen's terms.}
The model starts from a colored wave pattern and predicts how it bends, oscillates, and changes strength over time. Color identifies the wave's phase, while brightness shows its magnitude.

\begin{figure}[htbp]
  \centering
  \includegraphics[width=\linewidth,height=0.76\textheight,keepaspectratio]{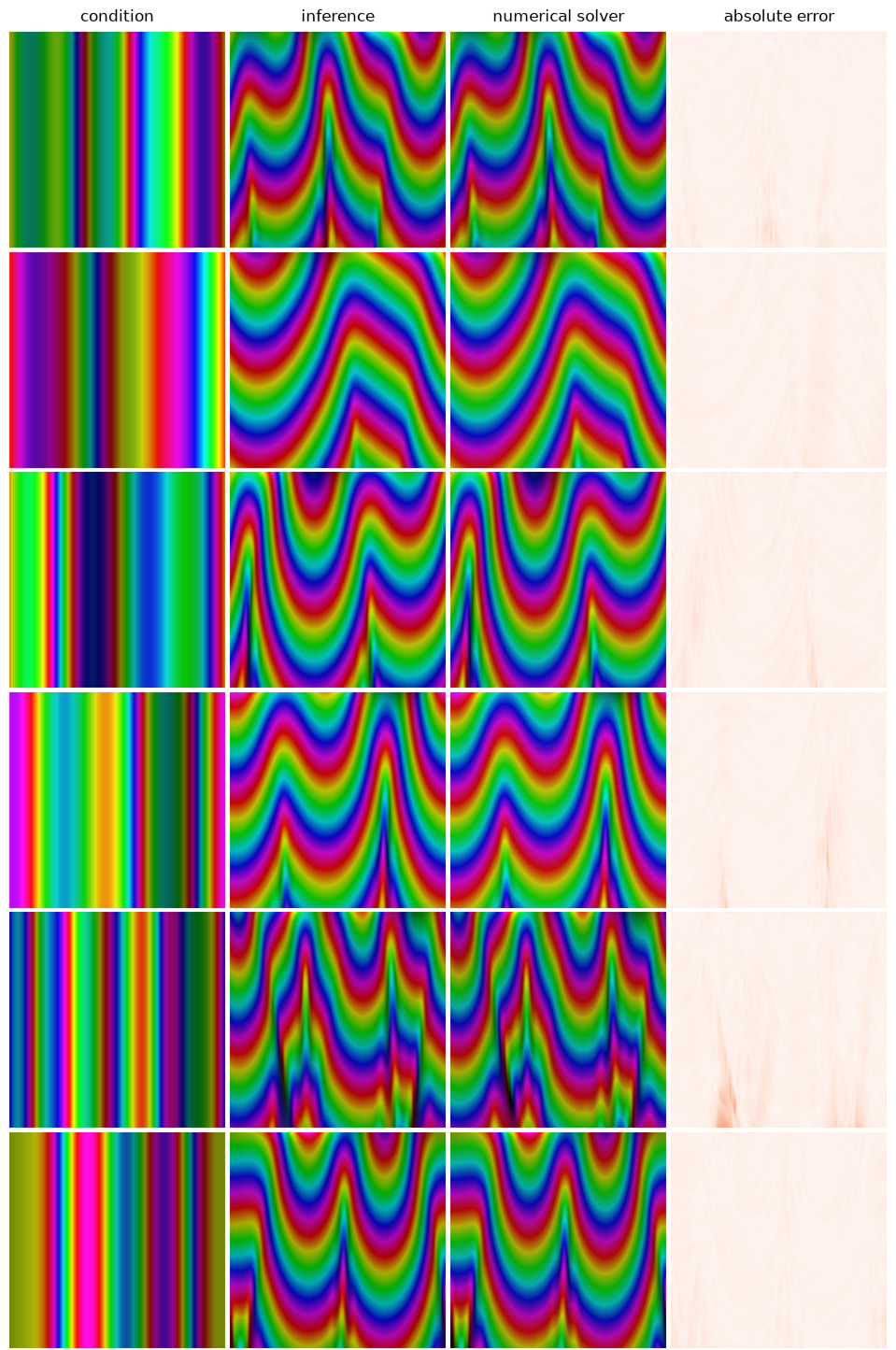}
  \caption{Random, non-handpicked held-out complex Ginzburg-Landau trajectories. Columns show the repeated initial condition, model inference, numerical space-time solution, and pointwise absolute error. Hue encodes complex phase, brightness encodes magnitude, and darker red marks larger error. Jagged paths reveal the instability of the dynamics.}
  \label{fig:cgl-results}
\end{figure}

\subsection{Shock-like dynamics: 1D Burgers equation}
\paragraph{Problem description.}
Burgers' equation \cite{burgers1948mathematical} evolves a one-dimensional field through nonlinear advection and diffusion. It is a canonical model of nonlinear steepening, viscous shock layers, and turbulence; related conservation-law formulations are also used in macroscopic traffic-flow models \cite{sharma2023burgerstraffic}. It tests whether the image editor can reproduce steepening and shock-like structures in a space-time image.

\paragraph{Equation formulation.}
For $x\in[0,1]$ and $t\in[0,T]$, the governing equation is
\begin{equation}
  \partial_t u(x,t) + u(x,t)\partial_x u(x,t) = \nu\partial_{xx}u(x,t).
\end{equation}
Here $u(x,t)$ is the scalar state, $\partial_x$ is the spatial derivative, $\partial_{xx}$ is the second spatial derivative, $\nu\geq 0$ is the diffusion or viscosity coefficient, and $T$ is the final simulated time.

\paragraph{Numerical solver method.}
WENO5 reconstruction handles nonlinear advection, third-order Runge-Kutta advances time, and centered differences approximate diffusion \cite{jiang1996weno,shu1988eno,gottlieb1998tvd}. Appendix~\ref{app:solver-burgers} gives the periodic boundary treatment, stability choices, resolution, and initial-condition sampler.

\paragraph{Image encoding.}
The image convention follows the heat-equation experiment: the horizontal axis is space and the vertical axis is time. The initial condition is encoded as a time-constant image, while the generated output is the time-varying solution field. The trajectory is area-downsampled to $256\times256$. Initial and target images share a symmetric per-sample color range given by the maximum absolute value of the trajectory.

\paragraph{In laymen's terms.}
The model watches an initial profile move and deform over time. Faster parts catch slower parts and form sharp fronts, while viscosity smooths those fronts.

\begin{figure}[htbp]
  \centering
  \includegraphics[width=\linewidth,height=0.76\textheight,keepaspectratio]{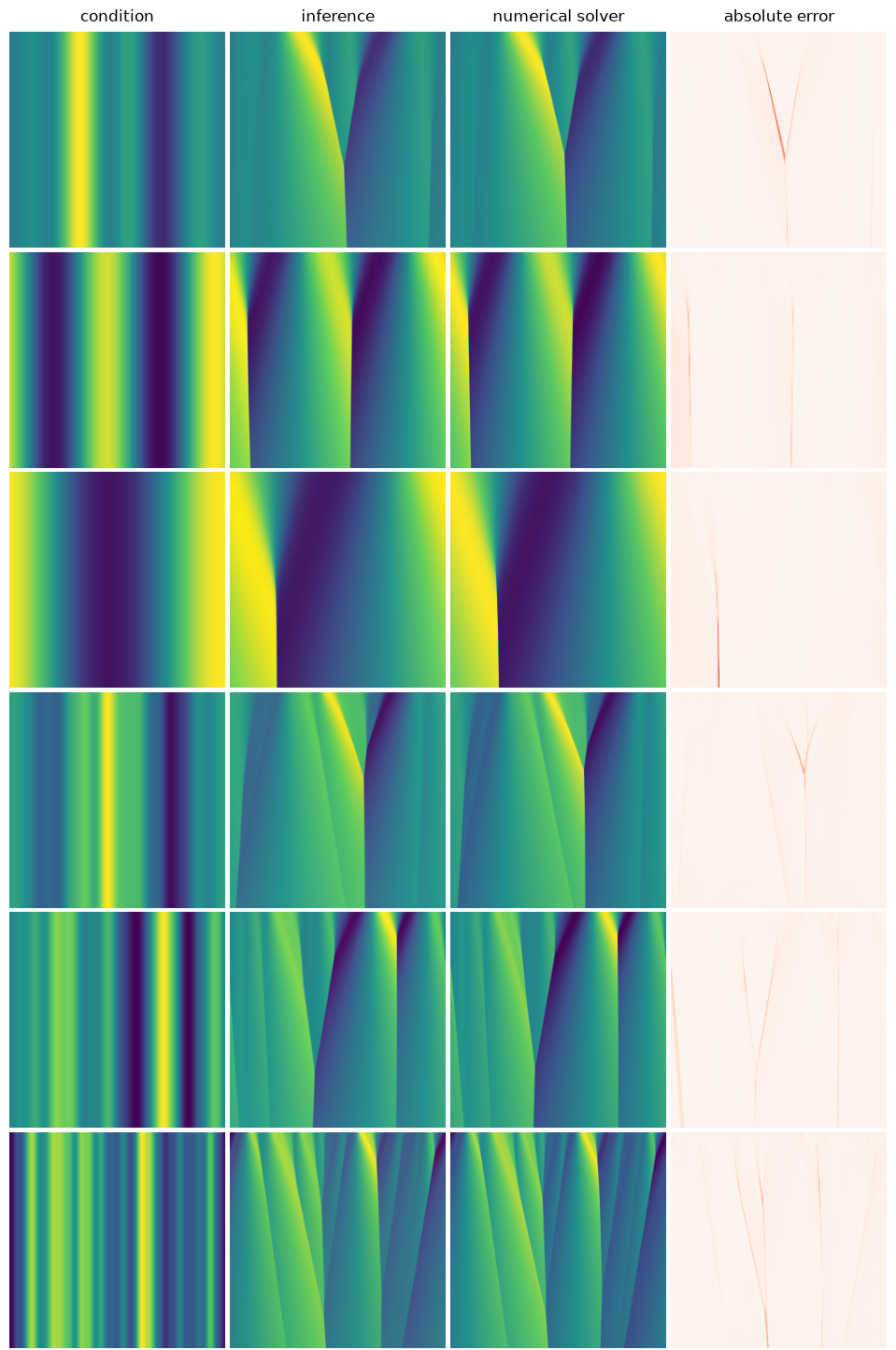}
  \caption{Random, non-handpicked held-out Burgers trajectories. Columns show the repeated initial condition, model inference, numerical space-time solution, and pointwise absolute error. Purple denotes lower and yellow higher field values on a shared per-sample scale; darker red denotes larger error. The sharp paths correspond to propagating shock-like fronts.}
  \label{fig:burgers-results}
\end{figure}

\subsection{2D future dynamics prediction: incompressible Navier-Stokes flow}
\paragraph{Problem description.}
The incompressible Navier-Stokes equations \cite{majda2002vorticity} describe the evolution of a velocity field under advection, pressure, and viscosity. This task predicts the two-dimensional field at $T=1$ from its initial state on a periodic unit square. Unlike the one-dimensional experiments, condition and target are complete spatial fields at two distinct times rather than one space-time diagram. No external forcing is included because a single initial image would not specify its values at intermediate times.

\paragraph{Equation formulation.}
For position $\mathbf{x}=(x,y)\in\Omega=[0,1]^2$ and time $t\in[0,T]$, the incompressible Navier-Stokes equations without forcing are
\begin{equation}
  \rho\left(\partial_t\mathbf{u}+(\mathbf{u}\cdot\nabla)\mathbf{u}\right)
  =-\nabla p+\rho\nu\Delta\mathbf{u},
  \qquad \nabla\cdot\mathbf{u}=0,
\end{equation}
The first equation balances fluid acceleration, pressure, and viscous diffusion; the second enforces local conservation of mass. Here $\mathbf{u}=(u_x,u_y)$ is velocity, $p$ is pressure, $\rho$ is fluid density, $\nu$ is kinematic viscosity, $\nabla$ is the spatial gradient, and $\Delta$ is the two-dimensional Laplacian. The corresponding dynamic viscosity is $\rho\nu$. All fields are periodic in both spatial directions and no external body force is applied. Taking the curl eliminates pressure and gives the scalar vorticity equation
\begin{equation}
  \partial_t\omega+\mathbf{u}\cdot\nabla\omega=\nu\Delta\omega,
  \qquad \omega=\partial_xu_y-\partial_yu_x,
\end{equation}
where $\omega$ is the out-of-plane vorticity.

\paragraph{Numerical solver method.}
The vorticity equation is integrated on a $128\times128$ Fourier pseudospectral grid \cite{peyret2002spectral}. The nonlinear term is evaluated in physical space with $2/3$ de-aliasing, viscous diffusion is applied exactly through an integrating factor, and four-stage Runge-Kutta advances advection. Appendix~\ref{app:solver-navier-stokes} details initial-condition sampling, timestep selection, and spectral velocity reconstruction.

\paragraph{Image encoding.}
The conditioning and target images use the same direct RGB encoding. For speed $s=\|\mathbf{u}\|_2$ and unit direction $\widehat{\mathbf{u}}=\mathbf{u}/s$, with zero direction assigned where $s=0$, each pixel is
\begin{equation}
  R=\frac{\widehat{u}_x+1}{2},\qquad
  G=\frac{\widehat{u}_y+1}{2},\qquad
  B=\min(s,1).
\end{equation}
Thus red and green encode the two normalized direction components, while blue encodes speed. Simulated fields are bilinearly resized to $256\times256$ before training.

\paragraph{Conditioning numerical parameters.}
The AdaLN-Zero conditioning values are density $\rho$ and kinematic viscosity $\nu$. Density is sampled uniformly from $[0.8,1.2]$ and transformed linearly; viscosity is sampled log-uniformly from $[5\times10^{-4},3\times10^{-3}]$ and log-transformed before normalization. Because the implemented unforced velocity equation is parameterized directly by kinematic viscosity, $\rho$ cancels from the dynamics and is a redundant conditioning variable at fixed $\nu$ in the current dataset.

\paragraph{In laymen's terms.}
The model receives a map of a fluid's initial speed and direction and predicts the flow one second later. Viscosity controls how quickly small swirls are smoothed away.

\begin{figure}[htbp]
  \centering
  \includegraphics[width=\linewidth,height=0.82\textheight,keepaspectratio]{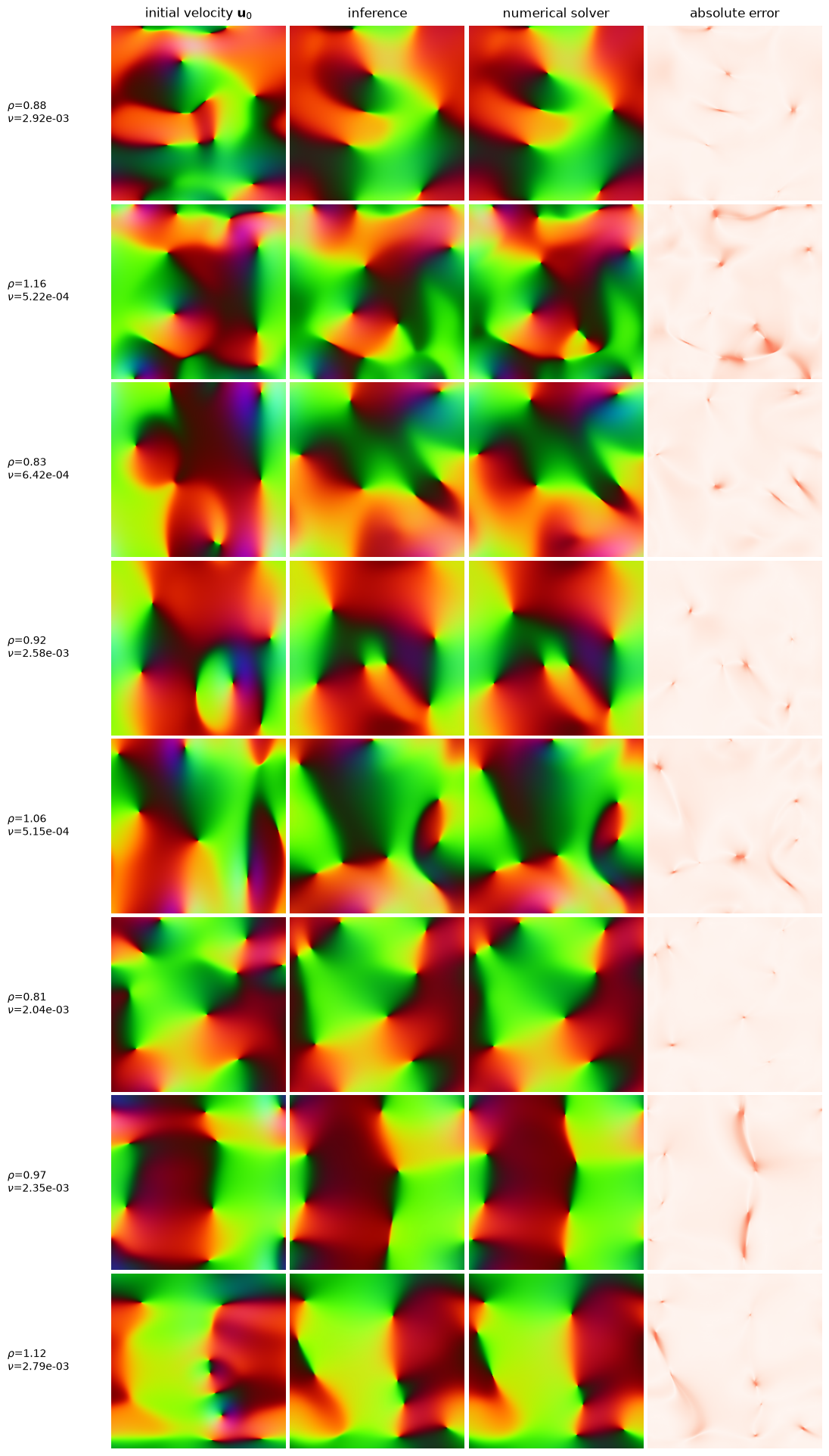}
  \caption{Random, non-handpicked held-out predictions for two-dimensional Navier-Stokes flow. Rows list density $\rho$ and kinematic viscosity $\nu$; columns show initial velocity $\mathbf{u}_0$, model inference, the pseudospectral solution at $T=1$, and absolute error. Red and green encode normalized horizontal and vertical direction, blue encodes speed, and darker red in the error map indicates a larger mean RGB discrepancy.}
  \label{fig:navier-stokes-results}
\end{figure}

\subsection{Potential-flow aerodynamics}
\paragraph{Problem description.}
Potential flow \cite{hess1967potential} approximates inviscid, incompressible, and irrotational motion around a body. This experiment predicts the static flow around a randomly generated two-dimensional shape. Each body is a closed, non-self-intersecting piecewise Bezier curve with 5-10 control points, scaled into the central half of the domain. Uniform far-field flow enters from the left with speed $U=1$.

\paragraph{Equation formulation.}
The simulation solves incompressible inviscid potential flow around the masked body using a stream function $\psi(x,y)$:
\begin{equation}
  \Delta \psi = 0, \qquad
  u = \partial_y \psi, \qquad
  v = -\partial_x \psi.
\end{equation}
Here $\Delta$ is the two-dimensional Laplacian, $\psi$ is the stream function, $u(x,y)$ and $v(x,y)$ are the horizontal and vertical velocity components, and $\partial_x,\partial_y$ are spatial derivatives. The outer box is assigned $\psi=Uy$, where $U$ is the incoming flow speed, and the body interior is assigned $\psi=U(n/2)$ on the $n\times n$ grid; generated bodies are centered at this height. Speed is recovered as $s(x,y)=\sqrt{u(x,y)^2+v(x,y)^2}$.

\paragraph{Numerical solver method.}
The stream-function Laplace equation is discretized with a five-point finite-difference stencil and solved directly as a sparse system on a $256\times256$ grid \cite{leveque2007finite}. Appendix~\ref{app:solver-potential-flow} details the body and outer-boundary treatment, velocity recovery, and modeling assumptions.

\paragraph{Image encoding.}
A rendered uniform-flow field, rather than a binary mask, supplies the body geometry: it places the body inside a uniform, unperturbed rightward flow using the same encoding as the target. In both images the body is white with a black outline. Flow direction selects a color from a cyclic six-color ramp after applying a phase gain of 5 and offset of 0.5; the resulting phase is clipped rather than wrapped. Speed controls brightness through a soft rolloff over the fixed interval $[0.5,2.1]$ followed by gamma correction with exponent 0.55. Thus RGB jointly represents direction and speed rather than storing physical components in separate channels. This encoding gives distant flow perturbations greater perceptual contrast than a direct linear mapping of the two velocity components.

\paragraph{In laymen's terms.}
The model sees an object placed in a steady wind and predicts how the air bends and changes speed around it. Color indicates direction and brightness indicates speed.

\begin{figure}[htbp]
  \centering
  \includegraphics[width=\linewidth,height=0.76\textheight,keepaspectratio]{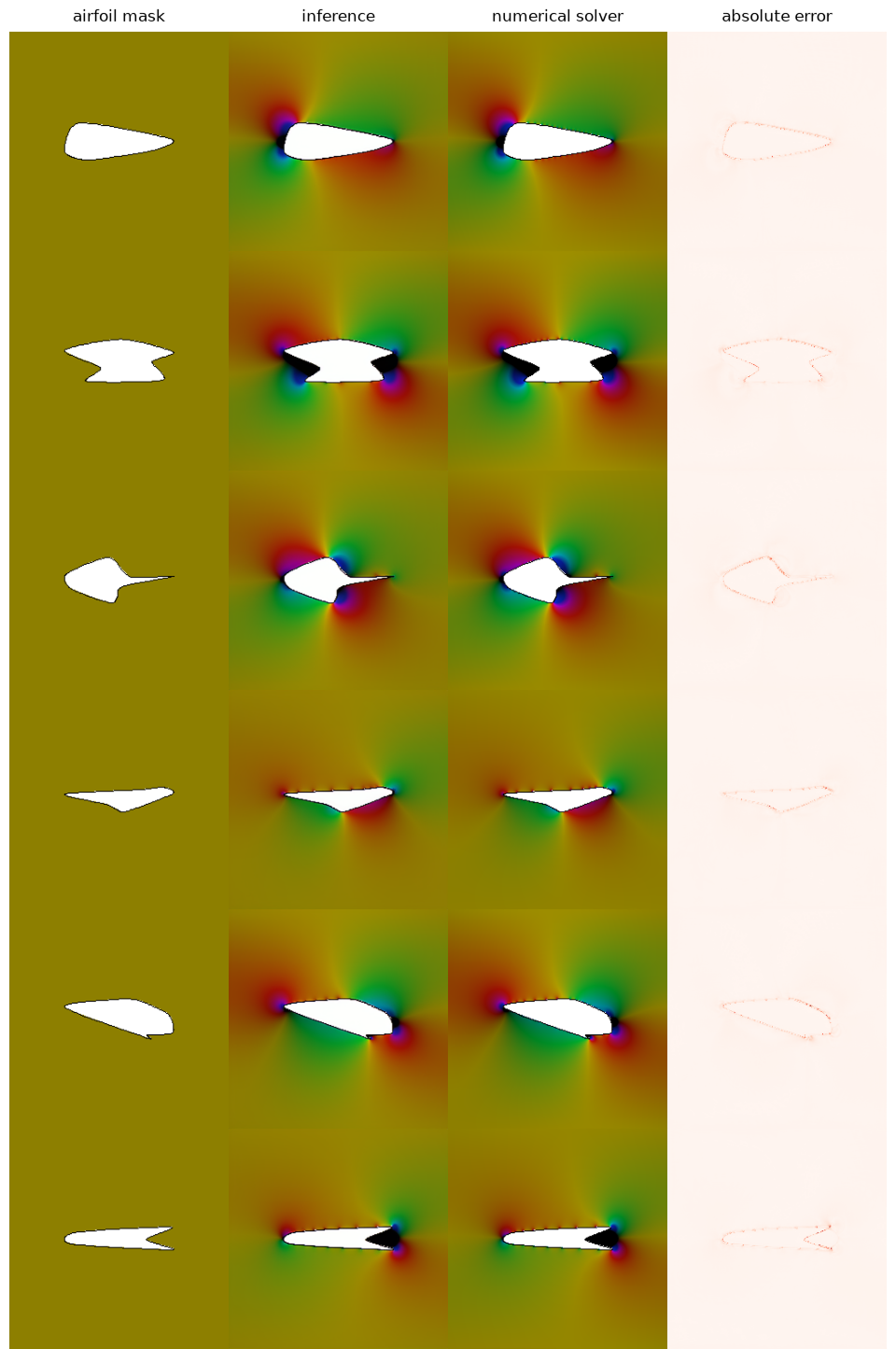}
  \caption{Random, non-handpicked held-out potential-flow samples. Columns show the uniform-flow condition with the embedded body, model inference, numerical solution, and pointwise absolute error. Hue encodes flow direction, brightness encodes speed, white marks the body, and darker red indicates larger error.}
  \label{fig:airfoil-results}
\end{figure}

\subsection{Many numerical conditionings: elastic stress around holes}
\paragraph{Problem description.}
Linear elasticity \cite{sadd2020elasticity} relates material deformation to applied loads. This experiment predicts stress concentration in a two-dimensional plate containing a random hole. Geometry varies as in the aerodynamics experiment, while material properties and far-field loads vary independently.

\paragraph{Equation formulation.}
The continuum model is linear elasticity:
\begin{equation}
  \nabla \cdot \sigma = 0, \qquad
  \sigma = C\epsilon, \qquad
  \epsilon = \frac{1}{2}\left(\nabla u + \nabla u^\top\right).
\end{equation}
Here $u(x,y)$ is the displacement vector, $\epsilon$ is the infinitesimal strain tensor, $\sigma$ is the stress tensor, and $C$ is the plane-stress constitutive matrix determined by the supplied Lamé parameters $\lambda$ and $\mu$. The implementation uses the effective plane-stress coefficient $\lambda_{\mathrm{ps}}=2\lambda\mu/(\lambda+2\mu)$ in $C$. For each triangular element $T$, the finite-element stiffness is
\begin{equation}
  K_e = |T| B^\top C B.
\end{equation}
In this expression, $K_e$ is the element stiffness matrix, $|T|$ is the triangle area, and $B$ is the constant strain-displacement matrix for the triangle.

\paragraph{Numerical solver method.}
Constant-strain triangular finite elements \cite{turner1956stiffness} discretize a $256\times256$ nodal grid, and a sparse direct solve recovers displacement. Appendix~\ref{app:solver-elasticity} describes mesh construction, the traction-free hole, biaxial boundary conditions, and matrix assembly.

\paragraph{Image encoding.}
A binary hole mask supplies the geometric condition. After displacement is solved, element stresses are averaged at grid nodes. Hue encodes principal-stress orientation, while brightness represents von Mises stress. Magnitude is clipped at the per-sample 99th percentile and gamma-corrected with exponent 0.55 to expose variations away from the hole.

\paragraph{Conditioning numerical parameters.}
The AdaLN-Zero scalar conditioning values for this method are the Lamé material parameters $\lambda$ and $\mu$ and the far-field loading values $\sigma_x$ and $\sigma_y$. Before normalization, the transform order is $\log \lambda$, $\log \mu$, linear $\sigma_x$, and linear $\sigma_y$. The material parameters are sampled log-uniformly from $[0.5,3]$, while each far-field stress is sampled uniformly from $[0.25,1.5]$. The random hole uses 3-10 Bezier control points and occupies at most approximately $100/256$ of the plate width.

\paragraph{In laymen's terms.}
The model receives the shape of a hole, the material stiffness, and the pulls applied from two directions. It predicts where stress concentrates and the orientation of that stress around the hole.

\begin{figure}[htbp]
  \centering
  \includegraphics[width=\linewidth,height=0.76\textheight,keepaspectratio]{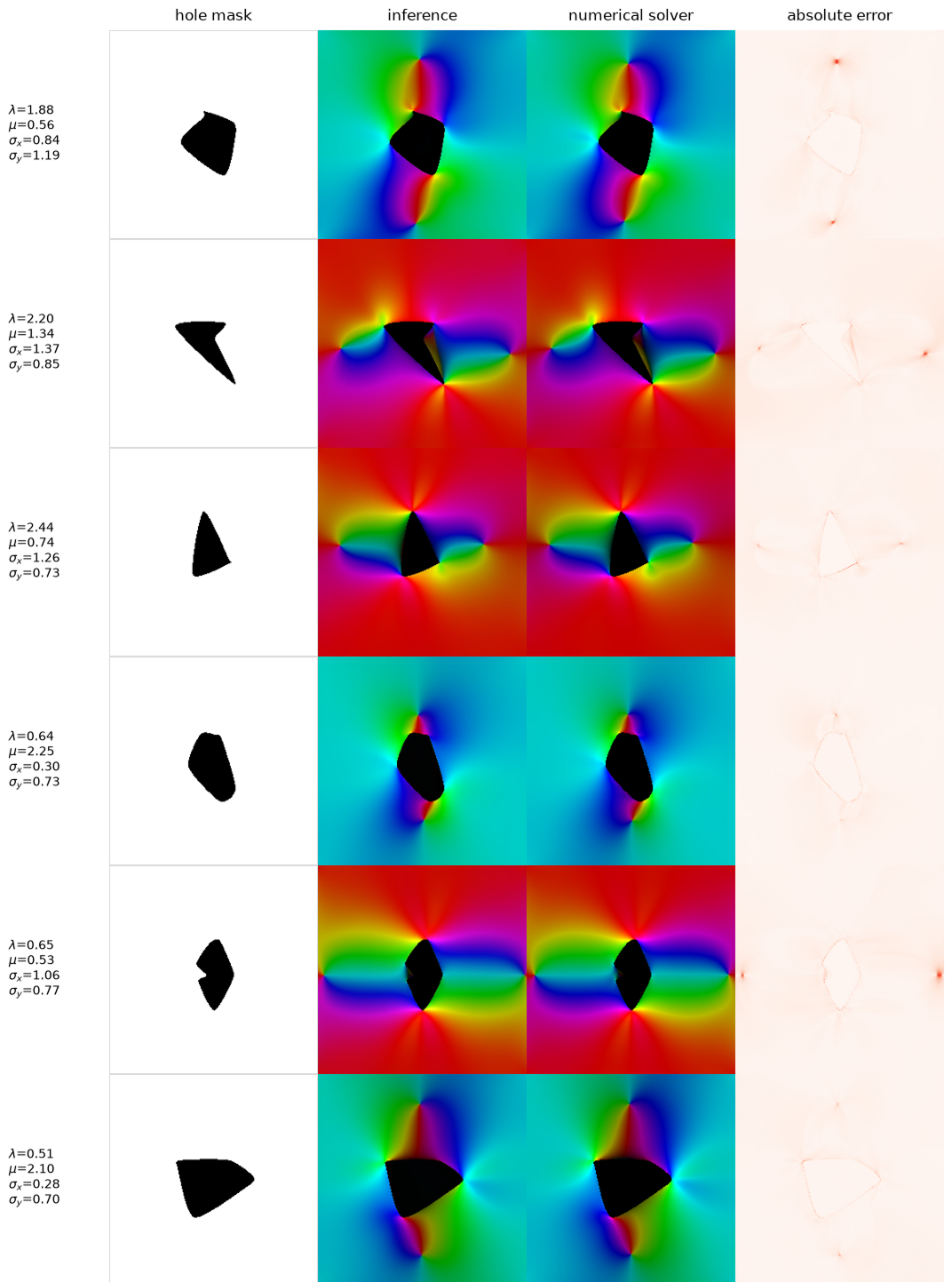}
  \caption{Random, non-handpicked held-out elastic-stress samples with varied holes, Lam\'e parameters $\lambda$ and $\mu$, and biaxial loads $\sigma_x$ and $\sigma_y$. Columns show the hole mask, model inference, finite-element solution, and pointwise absolute error. Hue encodes principal-stress orientation, brightness encodes von Mises magnitude, black marks the hole, and darker red denotes larger error.}
  \label{fig:elasticity-results}
\end{figure}

\subsection{Nonlinear first-arrival prediction: wave propagation in a heterogeneous medium}
\paragraph{Problem description.}
The eikonal equation \cite{nowack1992eikonal} gives the first-arrival time of a front traveling through a spatially varying medium. The model infers travel time from the center of a square through a refractive-index or slowness field. High-slowness regions delay propagation, while low-slowness regions permit faster travel. Related formulations describe light in heterogeneous media and shortest-time motion through spatially varying speed fields.

\paragraph{Equation formulation.}
The governing eikonal equation is
\begin{equation}
  |\nabla T(x,y)| = n(x,y), \qquad T(x_c,y_c)=0.
\end{equation}
Here $T(x,y)$ is the travel time from the source, $n(x,y)$ is the slowness field (rendered and labeled as refractive index), and $(x_c,y_c)$ is the center source location where travel time is zero.

\paragraph{Numerical solver method.}
The Fast Marching Method \cite{sethian1996fastmarching} propagates the front from the center on a $256\times256$ grid by repeatedly accepting the smallest tentative travel time and applying an upwind local update. Appendix~\ref{app:solver-eikonal} specifies the update, heap ordering, and random slowness fields.

\paragraph{Image encoding.}
The refractive-index field $n(x,y)$ serves as the condition and is sampled as a mixture of anisotropic Gaussian blobs. The output image represents the travel-time field $T(x,y)$ from the center source. Slowness uses the fixed range $[0.05,1]$; travel time uses $[0,0.75]$, gamma 0.6, and midpoint contrast 1.5. These fixed ranges cover the dataset while preserving visible color variation.

\paragraph{In laymen's terms.}
The model receives a map showing where motion is slow or fast and predicts how long a wave starting at the center takes to reach every location.

\begin{figure}[htbp]
  \centering
  \includegraphics[width=\linewidth,height=0.76\textheight,keepaspectratio]{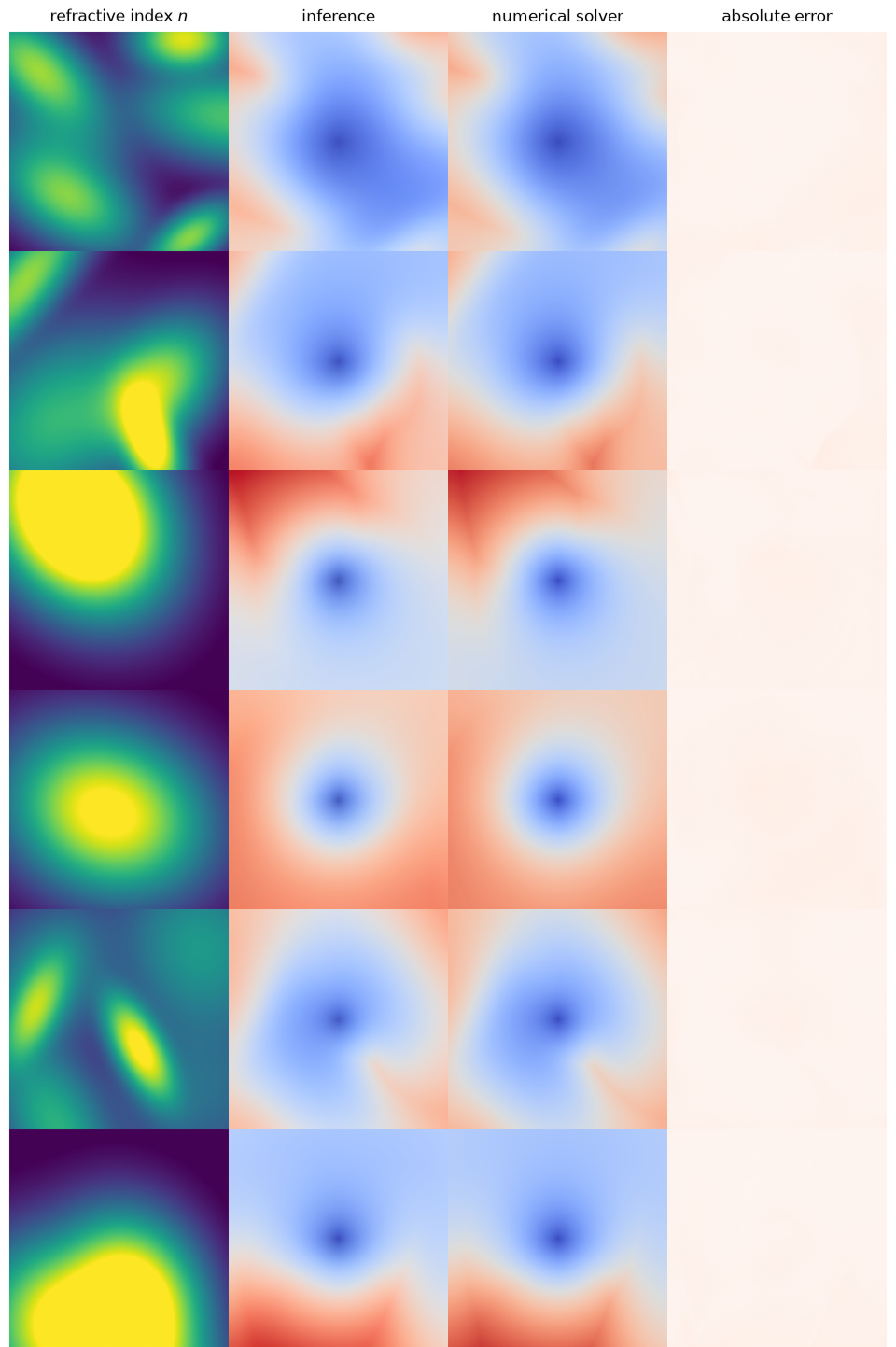}
  \caption{Random, non-handpicked held-out eikonal samples. Columns show refractive index $n$, model inference, the Fast Marching solution, and pointwise absolute error for travel time from the central source. In the input, purple-to-yellow maps low-to-high slowness; in the solution, blue marks short and red long travel times. Darker red in the error panel indicates a larger discrepancy.}
  \label{fig:eikonal-results}
\end{figure}

\subsection{Sparse conditions: crack damage propagation}
\paragraph{Problem description.}
Phase-field fracture \cite{miehe2010fracture} represents a crack as a continuous damage field coupled to elasticity. This task predicts damage growth from a boundary-originating crack under biaxial loading. The target is the final damage field rather than a binary crack path, so intermediate values describe partially degraded material.

\paragraph{Equation formulation.}
Damage is represented by a scalar field $d(x,y)\in[0,1]$, where $d=0$ denotes undamaged material and $d=1$ denotes fully damaged material. Elastic stiffness is degraded by
\begin{equation}
  g(d) = (1-d)^2 + \kappa,
\end{equation}
where $g(d)$ is the stiffness degradation factor and $\kappa>0$ is a small residual stiffness used to avoid singular stiffness matrices. The phase-field update solves a linearized damage equation of the form
\begin{equation}
  -G_c \ell \Delta d + \left(\frac{G_c}{\ell} + 2H\right)d = 2H,
\end{equation}
where $G_c$ is the critical fracture energy, $\ell$ is the phase-field length scale, $\Delta$ is the Laplacian, and $H(x,y)$ is the maximum total elastic strain-energy history accumulated over the loading sequence. This follows the history-field and staggered phase-field structure of Miehe et al. \cite{miehe2010fracture}, but our data generator uses total elastic strain energy rather than their tension-compression energy split. It is therefore a simplified surrogate formulation, not a reproduction of their full constitutive model.

\paragraph{Numerical solver method.}
A staggered finite-element solver alternates degraded-elasticity and damage updates while enforcing irreversibility. Appendix~\ref{app:solver-fracture} gives crack initialization, loading, regularization, and iteration details.

\paragraph{Image encoding.}
A rasterized initial crack field provides the image condition. The target image is the predicted final damage field $d(x,y)$ after the load history.

\paragraph{Conditioning numerical parameters.}
The AdaLN-Zero scalar conditioning values for this method are $\nu$, $G_c$, $\epsilon_h$, and $\epsilon_v$. Here $\nu$ is the elastic Poisson ratio for the fracture simulation, $G_c$ is the critical fracture energy, $\epsilon_h$ is the imposed horizontal strain, and $\epsilon_v$ is the imposed vertical strain. Before normalization, the transform order is linear $\nu$, $\log G_c$, linear $\epsilon_h$, and linear $\epsilon_v$. The phase-field length scale $\ell$ and residual stiffness $\kappa$ are solver constants in the current setup. Specifically, $\nu\sim U[0.18,0.35]$, $G_c$ is log-uniform on $[7\times10^{-5},2.2\times10^{-4}]$, $\ell=1.7/(128-1)$, and $\kappa=10^{-6}$. Loading is sampled from vertical-dominant, horizontal-dominant, and biaxial regimes; their low, high, and biaxial strain ranges are $[0.010,0.030]$, $[0.075,0.115]$, and $[0.055,0.090]$.

\paragraph{In laymen's terms.}
The model sees an initial crack together with material toughness and applied stretching, then predicts where the crack widens or grows. White denotes damaged material and black denotes intact material.

\begin{figure}[htbp]
  \centering
  \includegraphics[width=\linewidth,height=0.76\textheight,keepaspectratio]{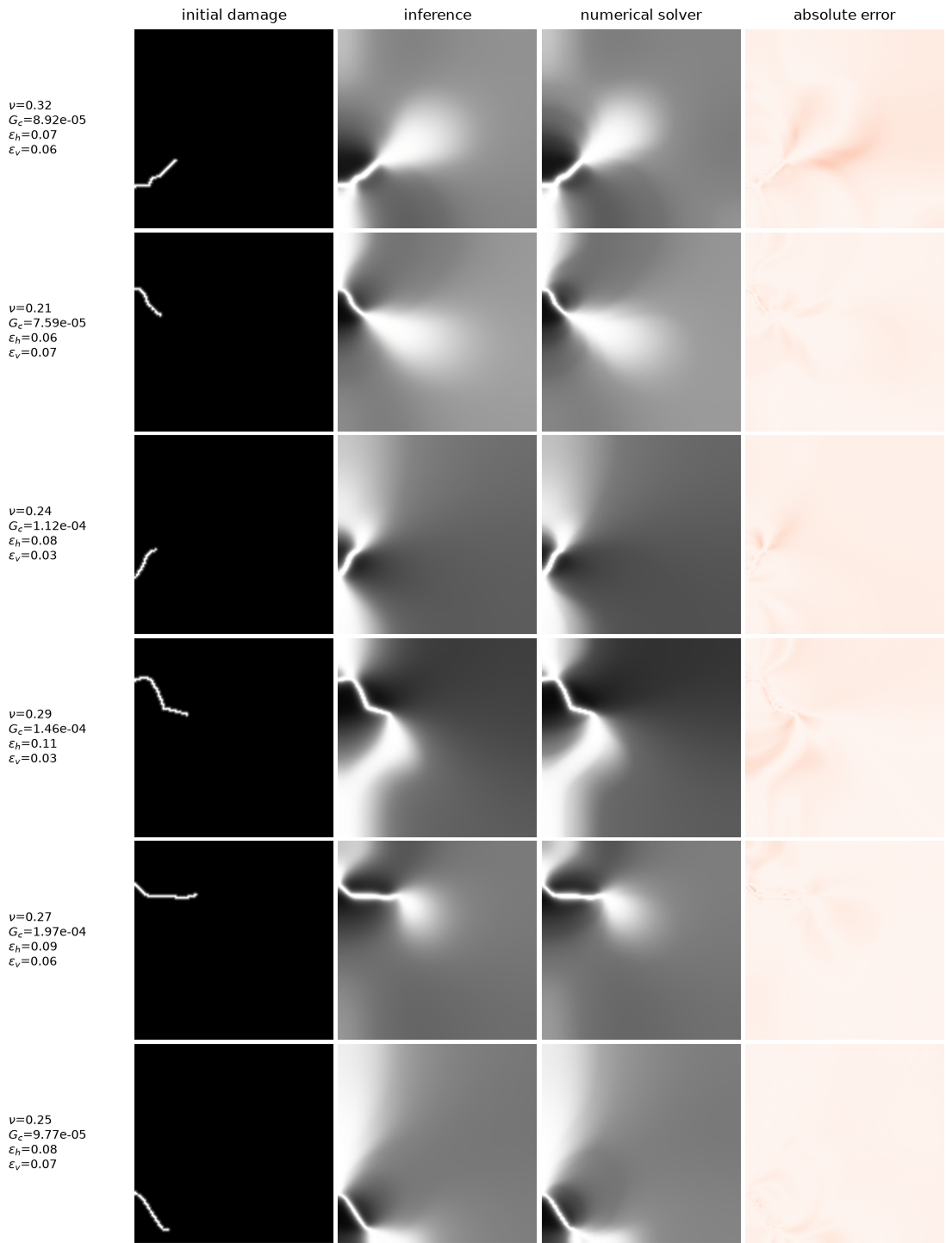}
  \caption{Random, non-handpicked held-out phase-field fracture samples with Poisson ratio $\nu$, fracture energy $G_c$, and imposed strains $\epsilon_h$ and $\epsilon_v$. Columns show initial damage, model inference, numerical solution, and pointwise absolute error. Grayscale maps intact material to black and increasing damage to white; darker red denotes larger error. Crack growth preferentially follows the direction of the larger applied strain.}
  \label{fig:fracture-results}
\end{figure}

\subsection{Entropic optimal transport}
\paragraph{Problem description.}
Entropic optimal transport and the Sinkhorn algorithm \cite{cuturi2013sinkhorn} approximate the least-cost rearrangement of one distribution into another using entropy-regularized matrix scaling. This experiment predicts a transport potential from source density, target density, and spatial cost. The densities are Gaussian mixtures. Unlike the preceding experiments, the target is defined by functional minimization rather than a PDE.

\paragraph{Equation formulation.}
For source grid point $x_i$ and target grid point $y_j$, the pairwise transport cost is the squared Euclidean distance modulated by the average cost along the straight path:
\begin{equation}
  \widetilde C_{ij} = \|x_i-y_j\|_2^2\left(1 + \rho \bar{c}_{ij}\right),
  \qquad C=\widetilde C/\max_{ij}\widetilde C_{ij}.
\end{equation}
Here $C_{ij}$ is the normalized cost of transporting mass from $x_i$ to $y_j$, $\rho$ is the cost-strength parameter, and $\bar{c}_{ij}$ is the average sampled cost field along the line segment between $x_i$ and $y_j$. With entropic regularization $\varepsilon>0$, the Sinkhorn kernel is
\begin{equation}
  K = \exp(-C/\varepsilon), \qquad
  r = a/(Ks), \qquad
  s = b/(K^\top r),
\end{equation}
where $a$ and $b$ are the source and target mass vectors, and $r$ and $s$ are the alternating Sinkhorn scaling vectors. The target image is the debiased source dual potential
\begin{equation}
  \phi =
  \varepsilon \log r_{\mathrm{source}\to\mathrm{target}}
  - \varepsilon \log r_{\mathrm{source}\to\mathrm{source}},
\end{equation}
where $\phi$ is the rendered potential, $r_{\mathrm{source}\to\mathrm{target}}$ is the source scaling from the source-to-target solve, and $r_{\mathrm{source}\to\mathrm{source}}$ is the source scaling from the debiasing source-to-source solve \cite{feydy2019sinkhorndivergences}.

\paragraph{Numerical solver method.}
We form the dense entropic kernel and use alternating Sinkhorn scaling \cite{cuturi2013sinkhorn}, followed by source-to-source debiasing \cite{feydy2019sinkhorndivergences}. We then mean-center the potential to fix its arbitrary additive constant. Grid size, path-cost sampling, regularization, and density generation are given in Appendix~\ref{app:solver-optimal-transport}.

\paragraph{Image encoding.}
Three images condition the model: log source density, log target density, and normalized spatial cost field. Densities use the fixed log range $[-33.6728,-9.1528]$, representative of the 1st-99th percentiles in the dataset. Log density places the conditioning channels on a variation scale closer to that of the transport potential. The output is the mean-centered debiased source dual potential, bicubically resized to $256\times256$ and rendered on the fixed signed range $[-0.4,0.4]$, representative of the dataset's extreme values.

\paragraph{In laymen's terms.}
The model receives a starting pile of mass, a desired final pile, and a map of where movement is expensive. It predicts a potential that indicates how the mass should be rearranged at minimum regularized cost.

\begin{figure}[htbp]
  \centering
  \includegraphics[width=\linewidth,height=0.76\textheight,keepaspectratio]{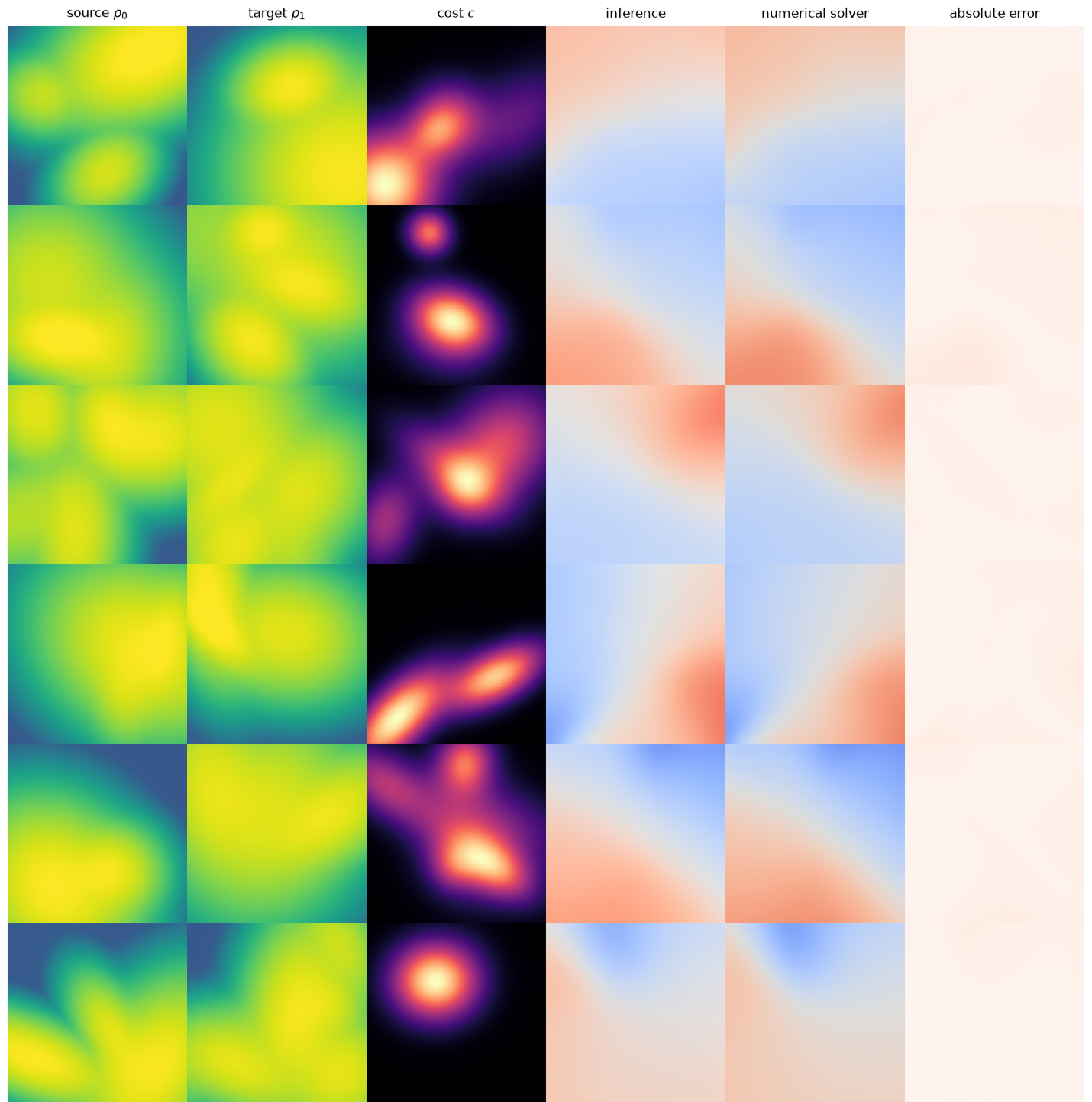}
  \caption{Random, non-handpicked held-out entropic optimal-transport samples. Columns show source log-density $\rho_0$, target log-density $\rho_1$, spatial cost $c$, model inference, Sinkhorn solution, and pointwise absolute error for the debiased source potential. Purple-to-yellow maps low-to-high log density, black-to-yellow maps low-to-high cost, blue-to-red maps negative-to-positive potential, and darker red denotes larger error.}
  \label{fig:optimal-transport-results}
\end{figure}

\FloatBarrier
\subsection{Other implementation details}
All experiments use $256\times256$ images and four-step inference. Rank-16 LoRA adapters and the scalar-conditioning projections are trained for 10,000 optimizer steps with an effective batch size of 4. Optimization uses AdamW \cite{loshchilov2019adamw} at an initial learning rate of $10^{-4}$; explicitly selected momentum coefficients and weight decay, together with the remaining PyTorch defaults, are reported in Appendix~\ref{app:model-implementation}. Frozen-VAE conditioning and target tokens are concatenated as the transformer's visual input. The appendix also specifies latent normalization, gradient clipping, and the scalar-conditioning network.

\section{Limitations}
This work demonstrates that an image-editing backbone can learn several solver-like mappings when physical states are rendered visually; it is not intended to outperform specialized numerical methods in accuracy, stability, or reliability. Conventional solvers still produce every target, and the learned model inherits their sampled data distribution together with the resolution and color-map choices of the image representation.

Image encoding also presumes that each physical channel can be assigned a useful numerical range across the training set. This is not always possible: quantities with inconsistent or unbounded sample-to-sample scales must either be clipped, losing information, or mapped into a shared interval that uses the available color range inefficiently. Although some of our experiments find success in per-sample normalization, generalizing this could alleviate this problem if the associated scale were supplied as an additional conditioning parameter. We leave the design and evaluation of such scale-aware encodings to future work.

The compressed latent representation introduces a second constraint. Because the model operates through a VAE \cite{kingma2014vae}, PDE residuals cannot be evaluated directly on its hidden state as they are in PINNs \cite{raissi2019pinn}; conservation constraints such as total mass or energy are similarly difficult to impose in a straightforward way. One possible direction is to learn latent-space surrogates for image-space physical losses and use them during generative training. It is unclear, however, whether such surrogates would remain meaningful for the off-manifold latent samples produced early in training, before the generator yields plausible decoded fields.

We also attempted to solve the one-dimensional Kuramoto-Sivashinsky equation (KS) \cite{cvitanovic2010ks}, a canonical model of spatiotemporal chaos. Positive Lyapunov exponents make chaotic trajectories exponentially sensitive to perturbations \cite{eckmann1985ergodic}, and we did not obtain satisfactory predictions. A major source of error was the round trip through the FLUX.2-klein VAE \cite{blackforestlabs2026flux2klein4b}, which produced reconstruction errors of up to approximately 2\% in initial-condition images. Since the initial field determines all subsequent dynamics, this perturbation changes the trajectory at its starting point and is then rapidly amplified. To isolate the effect, we encoded and decoded several initial-condition images, converted the reconstructions back to physical fields, and integrated those fields with the numerical simulator. Averaged over the tested conditions, the VAE round trip alone caused the trajectory error to reach 10\% by $t=5.9$ and explained approximately half of the total error observed at $t=50$. Extensive VAE fine-tuning delayed the 10\% threshold only to approximately $t=20$.

The remaining discrepancy is not confined to target reconstruction. FLUX updates every transformer token, including condition-image tokens, instead of preserving an exact representation of the supplied initial state throughout generation. Drift in those tokens therefore perturbs the condition that governs the predicted dynamics. We tried to reduce this mixing by reinjecting the condition representation into intermediate blocks through AdaLN-Zero modulation \cite{peebles2023dit} and dedicated cross-attention, without meaningful improvement. Adding LoRA \cite{hu2022lora} adapters to nearly every transformer layer was likewise ineffective. VAE distortion, condition-token drift, and chaotic amplification together leave too short a horizon for meaningful evolution. Without a substantially different representation or architecture, the present approach is therefore unlikely to simulate chaotic systems over useful timescales.

\paragraph{In laymen's terms.}
In a chaotic system, a tiny change to the starting image quickly produces a completely different future. Compression and generation both alter that starting information slightly, so even visually small errors become dominant after a short time.

\begin{figure}[htbp]
  \centering
  \includegraphics[width=0.92\linewidth]{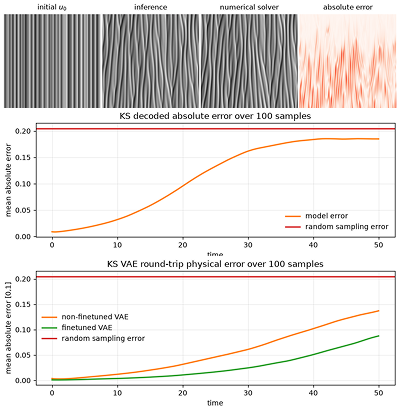}
  \caption{Random, non-handpicked held-out Kuramoto-Sivashinsky samples, including a model whose VAE was fine-tuned on KS data. Top: initial condition, model inference, numerical simulation, and absolute error; grayscale brightness encodes the normalized scalar field and darker red indicates larger error. Bottom: mean absolute error over simulated time, averaged across 10 samples. The numerical simulation is parameterized to remain non-divergent over the displayed horizon, yet model predictions rapidly diverge and reach the error of a randomly sampled solution near $t=30$.}
  \label{fig:ks-failure}
\end{figure}

\FloatBarrier
\section{Conclusions and Future Directions}
We have presented image editing as a common interface for physical simulation. Task-specific encoders render physical inputs as condition images, a pretrained generative backbone produces the corresponding solution fields, and lightweight adapters introduce scalar parameters when the images do not fully specify the problem. This procedure represented heterogeneous static equations, one- and two-dimensional dynamics, transport, and geometry-dependent mechanics within the same modeling framework.

The results establish capability rather than superiority over specialized solvers. Training depends on solver-generated examples, and the model offers no formal guarantees of accuracy, conservation, or stability. The failed Kuramoto-Sivashinsky experiment \cite{cvitanovic2010ks} makes this distinction concrete: VAE reconstruction error \cite{kingma2014vae} and condition-token drift are exponentially amplified in a system with positive Lyapunov exponents \cite{eckmann1985ergodic}, preventing useful long-horizon prediction.

At present, each equation requires its own fine-tuned adapter set. A broader system could instead share one model across related task classes through a consistent visual grammar, a textual specification of the governing PDE, and standardized channels for boundary conditions and physical parameters. PhysiX \cite{nguyen2025physix} reports that joint training across multiple physical tasks improves performance on individual tasks, suggesting that shared training could likewise strengthen this image-editing formulation while removing the need for one checkpoint per equation.

Another promising direction is solver-conditioned generation. A coarse numerical trajectory could serve as an input that the image model upsamples, corrects, or uses to reduce numerical drift when high-resolution integration is prohibitively expensive. This hybrid retains a conventional solver as an anchor and assigns the generative model a refinement role. It remains vulnerable in highly chaotic systems, where the coarse trajectory may cease to represent the desired dynamics before correction is possible.

\bibliographystyle{plain}
\bibliography{references}

\clearpage
\appendix
\section{Implementation and Simulator Details}

\subsection{Model adaptation and training}
\label{app:model-implementation}

\begin{wrapfigure}{r}{0.33\linewidth}
  \centering
  \includegraphics[width=\linewidth]{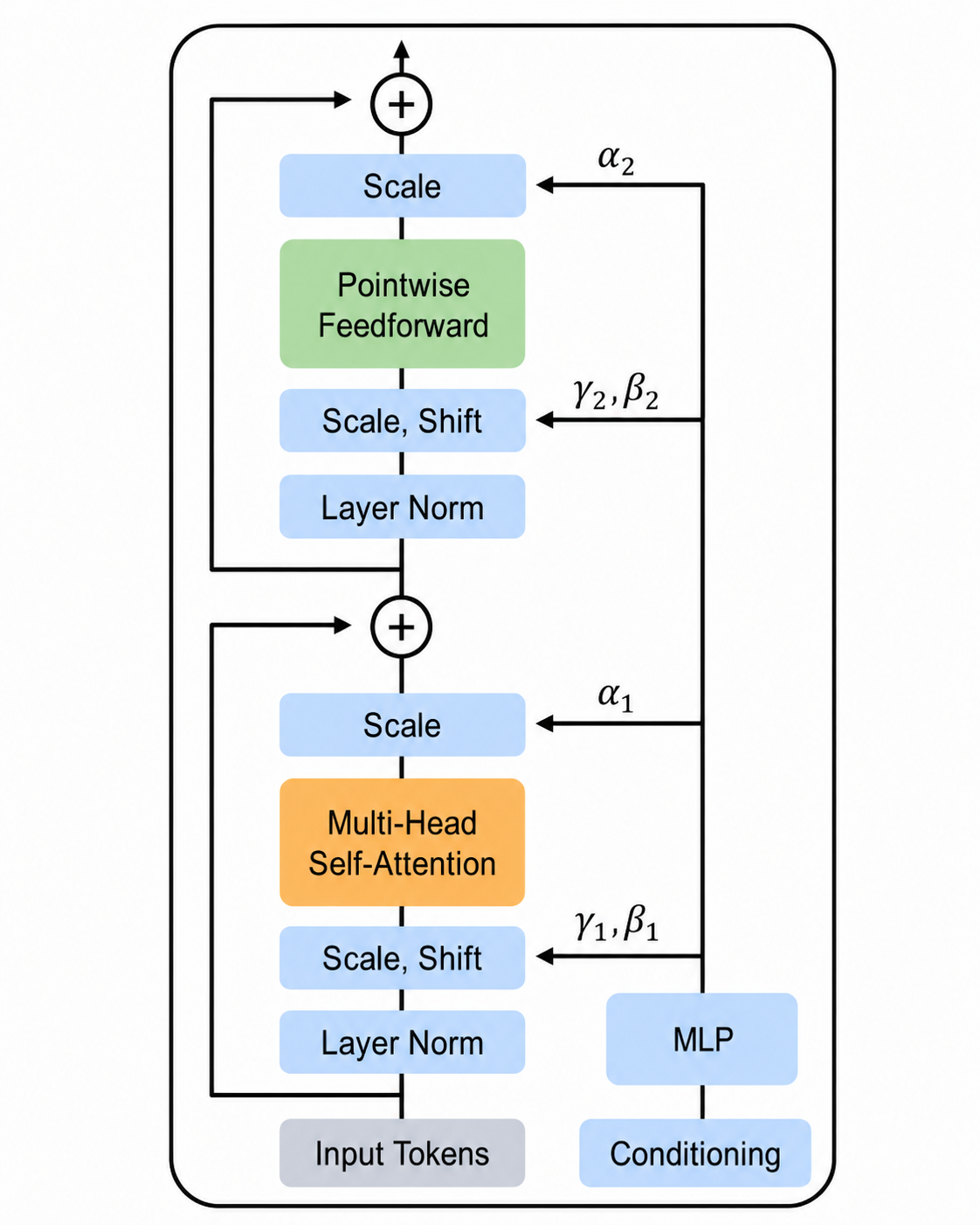}
  \caption{AdaLN-Zero modulation in a transformer block.}
  \label{fig:adaln-zero}
\end{wrapfigure}
AdaLN-Zero \cite{peebles2023dit} injects a normalized scalar-conditioning vector $z$ through six modulation vectors in each dual-stream block: attention shift $\beta_{\mathrm{attn}}$, scale $\gamma_{\mathrm{attn}}$, and gate $\alpha_{\mathrm{attn}}$, followed by the corresponding MLP shift, scale, and gate. In Figure~\ref{fig:adaln-zero}, these are $(\beta_1,\gamma_1,\alpha_1)$ and $(\beta_2,\gamma_2,\alpha_2)$. For block activation $h$,
\begin{equation}
\begin{aligned}
  \tilde h_{\mathrm{attn}} &=(1+\gamma_{\mathrm{attn}}(z))\odot
  \operatorname{LN}(h)+\beta_{\mathrm{attn}}(z),\\
  h' &=h+\alpha_{\mathrm{attn}}(z)\odot
  \operatorname{Attn}(\tilde h_{\mathrm{attn}}),\\
  \tilde h_{\mathrm{mlp}} &=(1+\gamma_{\mathrm{mlp}}(z))\odot
  \operatorname{LN}(h')+\beta_{\mathrm{mlp}}(z),\\
  h_{\mathrm{out}} &=h'+\alpha_{\mathrm{mlp}}(z)\odot
  \operatorname{MLP}(\tilde h_{\mathrm{mlp}}).
\end{aligned}
\end{equation}
Here $\operatorname{LN}$, $\operatorname{Attn}$, and $\operatorname{MLP}$ are layer normalization, self-attention, and the feed-forward sublayer, and $\odot$ is elementwise multiplication. The output projection is zero-initialized. In single-stream blocks, the two three-vector halves of the $6d_{\mathrm{DiT}}$ output are summed into one shift-scale-gate triplet.

For conditional flow matching \cite{lipman2023flowmatching}, let $x_1$ be the target VAE \cite{kingma2014vae} latent, $\epsilon\sim\mathcal{N}(0,I)$, and $c$ contain image and scalar conditions. At a selected schedule value $\sigma$,
\begin{equation}
  x_\sigma=(1-\sigma)x_1+\sigma\epsilon,
  \qquad u(x_\sigma\mid\epsilon,x_1)=\epsilon-x_1,
\end{equation}
and the objective is
\begin{equation}
  \mathcal{L}_{\mathrm{FM}}=\mathbb{E}_{\epsilon,x_1,\sigma}
  \left[\|v_\theta(x_\sigma,\sigma,c)-(\epsilon-x_1)\|_2^2\right].
\end{equation}
We uniformly select one of the four noise levels supplied by the distilled base model \cite{blackforestlabs2026flux2klein4b}; progressive distillation and consistency training provide general routes to few-step generation \cite{salimans2022progressive,song2023consistency}.

Training uses AdamW \cite{loshchilov2019adamw} with learning rate $10^{-4}$, $(\beta_1,\beta_2)=(0.9,0.999)$, weight decay $10^{-4}$, cosine decay, and gradient-norm clipping at 1.0. All other optimizer parameters use the PyTorch defaults. LoRA \cite{hu2022lora} rank and scaling are both 16. Normalized scalar parameters pass through a linear 64-dimensional projection, SiLU, and a $6d_{\mathrm{DiT}}$ projection. Microbatches of 2 and two-step accumulation give an effective batch size of 4; training lasts 10,000 optimizer steps. Images are encoded with the frozen VAE posterior mode, patchified, normalized with stored VAE batch-normalization statistics, and concatenated as noisy target and conditioning tokens.

\subsection{Two-dimensional elliptic solver}
\label{app:solver-elliptic}
Centered differences \cite{narasimhan2008laplace} approximate first and unmixed second derivatives on a $256\times256$ grid; the mixed derivative uses four diagonal neighbors, giving a nine-point stencil. The sparse system is solved directly. Coefficient fields are low-pass-filtered white noise with at most 5.5 spatial cycles. We sample $a_{20},a_{02}\in[0.015,0.12]$ and set $a_{11}=2\rho\sqrt{a_{20}a_{02}}\eta$ with $\rho=0.35$ and $\eta\in[-1,1]$. First-order coefficients lie in $[-0.08,0.08]$; for $q\in[0.5,10]$ and $r\in[0,1]$, $a_{00}=-q$ and $f=a_{00}r$.

\subsection{Forced heat-equation solver}
\label{app:solver-heat}
The periodic field is expanded in Fourier modes \cite{leveque2007finite}, where the Laplacian is diagonal. Diffusion is advanced by its exact heat multiplier and the linearly time-interpolated forcing is integrated analytically in the same basis. We evaluate 512 times to $T=1.5$ on 2048 spatial points before area downsampling. Initial conditions are
\begin{equation}
  u_0(x)=\sum_{k=1}^{M}a_k\sin(2\pi kx+\phi_k),
\end{equation}
where $M$ is uniform on 1-24, Gaussian amplitudes decay as $1/k$, and the signal is normalized to unit standard deviation. Forcing endpoints use 1-12 independently sampled active modes.

\subsection{Complex Ginzburg-Landau solver}
\label{app:solver-cgl}
With Fourier wavenumber $k$ and substep $\Delta t$, define
\begin{equation}
 L(k)=1-(1+ic_1)k^2,\quad E(k)=e^{L(k)\Delta t},\quad
 \varphi(k)=\frac{E(k)-1}{L(k)},
\end{equation}
and $N(A)=-(1-ic_3)|A|^2A$. Exponential Euler \cite{cox2002etd} updates
\begin{equation}
 A^{n+1}=\mathcal{F}^{-1}\left[E(k)\widehat{A^n}(k)
 +\varphi(k)\widehat{N(A^n)}(k)\right],
\end{equation}
integrating the stiff linear term exactly and the cubic term explicitly. We use 2048 spatial points, 512 stored frames, and eight substeps per frame. Independent phase and amplitude Fourier series use 2-8 active modes with coefficients decaying as $1/k$. Phase has standard deviation $\pi$; amplitude is $\max(1+0.25\xi(x),0.05)$ for unit-variance $\xi$.

\subsection{Burgers solver}
\label{app:solver-burgers}
WENO5 combines candidate advective-flux reconstructions with nonlinear smoothness weights, while a global Lax-Friedrichs split separates propagation directions \cite{jiang1996weno,shu1988eno}. Diffusion uses centered differences, and third-order TVD Runge-Kutta advances time \cite{gottlieb1998tvd}. All stencils are periodic. The fixed step is the minimum of initial advective CFL and diffusive stability bounds. We use $T=0.15$, $\nu=3\times10^{-5}$, 2048 spatial points, and 512 stored times. Initial conditions contain 1-24 Fourier modes with Gaussian $1/k$ coefficients, random phases, and standard deviation 1.5.

\subsection{Two-dimensional Navier-Stokes solver}
\label{app:solver-navier-stokes}
We solve the periodic vorticity equation \cite{majda2002vorticity} on the unit square with $N=128$ points per direction. Initial conditions are generated from a periodic streamfunction formed by 3-8 wrapped anisotropic Gaussian components. Component centers are uniform on the domain, each width is sampled independently from $[0.055,0.16]$, and component weights are standard Gaussian. After subtracting the mean streamfunction, we set
\begin{equation}
  \mathbf{u}_0=(\partial_y\psi_0,-\partial_x\psi_0),
\end{equation}
which makes the initial velocity divergence-free, and rescale it so that its maximum speed is uniform on $[0.35,0.75]$.

At each step, a Fourier pseudospectral method \cite{peyret2002spectral} recovers velocity from vorticity through the streamfunction. The nonlinear term $-\mathbf{u}\cdot\nabla\omega$ is evaluated pseudospectrally, and modes with either integer wavenumber above $N/3$ are discarded. A four-stage integrating-factor Runge-Kutta update applies the diffusive multiplier $\exp(-\nu|\mathbf{k}|^2\Delta t)$ exactly and advances advection explicitly, where $\mathbf{k}$ is the Fourier wavevector and $\Delta t$ is the timestep. The step satisfies $\Delta t\leq0.4/N$; consequently, the integration from $t=0$ to $T=1$ uses 320 equal steps. The initial and final velocities are bilinearly resized from $128\times128$ to $256\times256$ before RGB encoding.

\subsection{Potential-flow solver}
\label{app:solver-potential-flow}
A five-point Laplacian is assembled over fluid cells on a $256\times256$ grid and solved directly \cite{leveque2007finite}. A constant stream function on the body enforces impermeability; the outer boundary prescribes uniform flow. Velocity is recovered by differentiating the solved stream function. This harmonic model captures inviscid irrotational flow but excludes viscous boundary layers and separated wakes \cite{hess1967potential}.

\subsection{Elasticity solver}
\label{app:solver-elasticity}
Each solid square cell in a $256\times256$ nodal grid is split into two constant-strain triangles; cells centered inside the hole are omitted. This omission leaves its boundary traction-free. Sampled far-field stresses are converted to plane-stress strains and imposed as affine outer-boundary displacements. Element matrices $|T|B^\top C B$ are assembled globally, fixed and free degrees of freedom are partitioned, and the sparse displacement system is solved directly \cite{turner1956stiffness}.

\subsection{Eikonal solver}
\label{app:solver-eikonal}
First-order fast marching maintains accepted and tentative nodes in a heap. The smallest tentative travel time is accepted, and neighboring values are relaxed with the upwind quadratic eikonal update using accepted horizontal and vertical neighbors \cite{sethian1996fastmarching}. We solve on a $256\times256$ grid. Slowness fields combine 1-12 oriented anisotropic Gaussians with widths in $[0.05,0.25]$ and lognormal weights, then are normalized to mean 0.5, contrast-adjusted, and clipped to $[0.05,1]$.

\subsection{Phase-field fracture solver}
\label{app:solver-fracture}
One piecewise-linear boundary-originating crack is rasterized as a narrow Gaussian on a $128\times128$ triangular mesh. Following the staggered history-field strategy of Miehe et al. \cite{miehe2010fracture}, each of four load steps alternates at most twice between degraded linear elasticity and the linear damage equation. Our simplified implementation uses total strain energy rather than the cited formulation's tension-compression split. Damage is clipped to enforce irreversibility and remain above its initial value. Biaxial displacements impose $\epsilon_h$ on left/right and $\epsilon_v$ on bottom/top boundaries. We use $E=1$ and multiply each element stiffness by $g(\bar d_e)$, where $\bar d_e$ is mean element damage.

\subsection{Entropic optimal-transport solver}
\label{app:solver-optimal-transport}
The dense cost matrix defines $K=\exp(-C/\varepsilon)$; alternating row and column scaling updates drive the coupling toward the source and target marginals \cite{cuturi2013sinkhorn}. We subtract the source-to-source potential and mean-center the result to reduce entropic smoothing bias \cite{feydy2019sinkhorndivergences}. The solve uses a $24\times24$ grid, $\rho=8$, nine path samples, $\varepsilon=0.055$, and 90 iterations. Source and target densities contain 2-6 Gaussian components with widths in $[0.05,0.18]$; cost-field widths are 1.5 times larger.

\end{document}